\documentclass{ieeeconf}
\usepackage{graphicx}
\usepackage{xcolor}
\usepackage{hyperref}
\usepackage{biblatex}
\usepackage{textcomp}
\usepackage{amsmath, amssymb}

\usepackage{tikz}
\usetikzlibrary{calc}

\newcommand{\SE}{\mathsf{SE}}
\newcommand{\Conf}{\mathcal{Q}}
\newcommand{\se}{\mathfrak{se}}

\begin{document}

\title{Bridging Walking and Slithering -- Stokesian Locomotion}
\author{Shai Revzen, Dan Zhao, Glenna Clifton, Brian Bittner, Nick Gravish}
\maketitle

\begin{tikzpicture}[remember picture, overlay]
    \draw[black, fill=black!20, fill opacity=0.7]
          ($(current page.north)+(-5.5,-1.5)$) circle (0cm)
          node {Presented at Dynamic Walking on May 20th 2021};
\end{tikzpicture}

\section{Introduction}

Both legged locomotion and slithering motions typically utilize periodic gaits -- repeating cycles of body shape change that produce a net motion through the world.
Legged locomotion can be viewed from the perspective of piece-wise contact constraint formation and removal.
Slithering and low Reynolds number swimming operate under continuous constraints of force balance, wherein dissipation removes the ability to accumulate momentum.
Here we discuss how to bridge the gap between these domains of motion, thereby, among other benefits, producing models for the space of legged locomotion \emph{with slipping}.
The connective fabric is the use of a ``Stokesian'', or ``local connection'' model of the form $g^{-1}\dot g= A(r) \dot r$ where $g \in \SE(3)$ is the body frame location and $r \in \Conf$ is the body shape.

\section{Holonomic Motion}
Leading up to locomotion, examine the ability of a robotic manipulator to move and orient its own center of mass\footnote{%
    We tacitly assume a CoM attached ``body frame''; the proper selection of such a frame is non-trivial \cite{hatton2011geometric}.}.
There is typically a map $F:\Conf \to \SE(3)$ from the configuration parameters describing the shape of the robot $r \in \Conf$, to the body frame in $g = F(r) \in \SE(3)$.
Here the constraints are ``holonomic'', i.e. they arise from the ``whole'' function $F$, and as a consequence no periodic motion in $r$ can lead to a net motion through the world.
When $r$ returns to its initial value, the $g$ also returns.
The constraint in this case can be studied using the Jacobian $J_F$ of $F$.
We can construct a body velocity map as a ``local connection'' 
$A_F(r) := (F(r))^{-1} J_F(r)$ and thus $A_F(r) \dot r \in \se(3)$ is a body velocity\footnote{%
   We assumed notation of a matrix Lie group representation for $\SE(3)$.
}.

\section{Piece-wise holonomic motion}
Typically, legged locomotion is modeled assuming non-slip contacts with the environment.
For simplicity, we restrict our treatment to sets of contact constraints that fully determine the body frame location relative to the feet, i.e. are not under-actuated.
These could be three or more non-colinear pin-joint foot contacts, two foot contacts at least one of which can bear a moment around the line connecting them, or one foot which can bear all three moments.

It is important to note that even for a simple case where which feet are in contact is uniquely determined by the body shape, $g$ can no longer be written as a function of $r$.
For every continuous set of contacts $c\in C$ out of a finite set of contact combinations $C$, there exists an $F[c]:\Conf \to \SE(3)$.
If any one of the contact combinations is maintained no motion is possible for the same reason above.
However motion is possible, because the change in contacts involves a jump discontinuity in $g$ -- the $F[c]$ do not meet continuously at the switching of $c$, i.e. in general $F[c_1](r)\neq F[c_2](r)$ even for $r$ values where $c_1$ can switch to $c_2$.

Here the local connection allows for a concise and precise representation using $A_{F[c]}(r)$.
Like $F[c]$ this function has a jump discontinuity when $c$ changes,
but it allow us to model motion using the ODE $g^{-1} \dot g = A_{F[c]}(r) \dot r$, which produces continuous solutions in $g$ -- the body frame.

\section{Slithering Motion}
For slithering and low Reynolds number (so called ``Stokes Fluid'') swimming, it is well established that a similar mathematical formalism $g^{-1} \dot g = A(r) \dot r$ applies.
However here $A(r)$ is a fundamental object, and nowhere is it derived from an underlying Jacobian.
This expresses itself in the fact that most Stokesian systems will move the body frame with almost every cyclic shape change, unlike piecewise holonomic systems which require a cycle that spans multiple of the individually holonomic pieces.
$A(r)$ could be approximated piece-wise with pieces that come from holonomic systems; in this sense slithering could arise as a limit of walking with an infinite number of legs.

\section{Walking with slipping}
Consider the removal of the non-slip requirements that typically produce holonomic motion.
With slipping $A(r)$ for a fixed set of contacts $c$ is no longer derived from a Jacobian of some underlying $F[c]$.
Instead, it is derived from a set of implicit equations tying together body frame velocity and shape change speed $\dot r$, i.e. $G(r ; g^{-1}\dot g, \dot r)=0$
If the $G$ is linear in $\dot r$ and in $g^{-1}\dot g$, the resulting system also takes the form $g^{-1} \dot g = A(r) \dot r$. 

\section{Summary}
The common mathematical formalism $g^{-1} \dot g = A(r) \dot r$ extends across all four types of motion -- holonomic, piece-wise holonomic, slithering, and walking with slipping.
It is the special case of the ``reconstruction equation'' of \cite{bloch2005nonholonomic} for the case where momentum can be ignored (see also \cite{Kvalheim2019gmpsr}).
\printbibliography
\end{document}